\documentclass{llncs}
\usepackage{makeidx}  
\usepackage{graphicx}
\usepackage{subcaption}
\usepackage{caption}
\usepackage{booktabs}
\usepackage{amssymb}
\usepackage[misc]{ifsym}
\begin{document}
\frontmatter          
\pagestyle{headings}  
%

\mainmatter              
\title{Liver Lesion Detection from Weakly-labeled Multi-phase CT Volumes with a Grouped Single Shot MultiBox Detector}
\titlerunning{Hamiltonian Mechanics}  
%
\author{Sang-gil Lee\inst{1}\and
Jae Seok Bae\inst{2,3}\and
Hyunjae Kim\inst{1}\and
Jung Hoon Kim\inst{2,3,4}\and
Sungroh Yoon\inst{1}\textsuperscript{(\Letter)}}
\authorrunning{Lee et al.} 
%
\tocauthor{}
\institute{Electrical and Computer Engineering, Seoul National University, Seoul, Korea\\
\email{sryoon@snu.ac.kr}\\
\and
Radiology, Seoul National University Hospital, Seoul, Korea\\
\and
Radiology, Seoul National University College of Medicine, Seoul, Korea\\
\and
Institute of Radiation Medicine, Seoul National University Medical Research Center, Seoul, Korea
}

\maketitle              

\begin{abstract}
We present a focal liver lesion detection model leveraged by custom-designed multi-phase computed tomography (CT) volumes, which reflects real-world clinical lesion detection practice using a Single Shot MultiBox Detector (SSD). We show that grouped convolutions effectively harness richer information of the multi-phase data for the object detection model, while a naive application of SSD suffers from a generalization gap. We trained and evaluated the modified SSD model and recently proposed variants with our CT dataset of 64 subjects by five-fold cross validation. Our model achieved a 53.3\% average precision score and ran in under three seconds per volume, outperforming the original model and state-of-the-art variants. Results show that the one-stage object detection model is a practical solution, which runs in near real-time and can learn an unbiased feature representation from a large-volume real-world detection dataset, which requires less tedious and time consuming construction of the weak phase-level bounding box labels.

\end{abstract}
\section{Introduction}
\begin{figure}[t]
	\centering
	\includegraphics[width=0.65\linewidth]{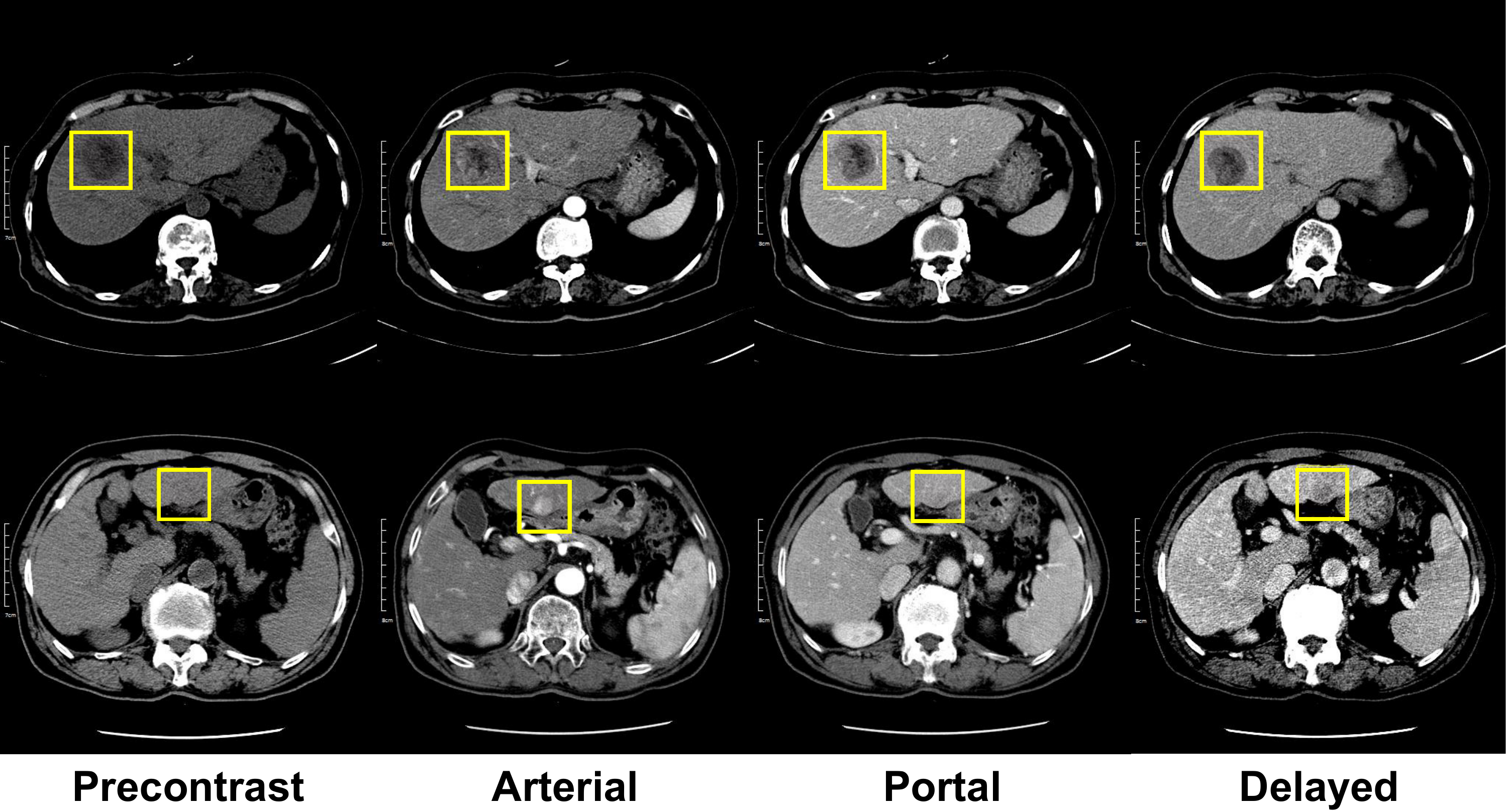}
    \caption{Examples of the multi-phase CT dataset. Top: Lesions are visible from all phases. Bottom: Specific variants of lesions are visible only from specific phases. Note that the lesions are barely visible from the portal phase.}
    \label{fig:figure1}
\end{figure}

Liver cancer is the sixth most common cancer in the world and the second most common cause of cancer-related mortality with an estimated 746,000 deaths worldwide per year \cite{cancerreport}. Of all primary liver cancers, hepatocellular carcinoma (HCC) represents approximately 80\% and most HCCs develop in patients with chronic liver disease \cite{hcc}. Furthermore, early diagnosis and treatment of HCC is known to yield better prognosis \cite{earlyhcc}. Therefore, it is of critical importance to be able to detect focal liver lesions in patients with chronic liver disease.
\\\\
\indent Among the various imaging modalities, computed tomography (CT) is the most widely utilized tool for HCC surveillance owing to its high diagnostic performance and excellent availability. A dynamic CT protocol of the liver consists of multiple phases \cite{multict}, including precontrast, arterial, portal, and delayed phases to aid in the detection of the HCCs that have different hemodynamics from surrounding normal liver parenchyma. However, as a result, dynamic CT of the liver produces a large number of images, which require much time and effort for radiologists to interpret. In addition, early stage HCCs tend to be indistinct or small and sometimes it is difficult to distinguish them from adjacent hepatic vasculatures or benign lesions, such as arterioportal shunts, hemangioma, etc. Hence, diagnostic performance for early stage HCCs using CT is low compared to large, overt HCCs \cite{earlyhcc2}. If focal liver lesions could be automatically pre-detected from CT images, radiologists would be able to avoid the laborious work of reading all images and focus only on the characterization of the focal liver lesions. Consequently, interpretation of liver CT images would be more efficient and expectedly also more accurate owing to focused reading.
\\\\
\indent Most publicly available CT datasets contain only the portal phase with per-pixel segmentation labeling \cite{3dircadb,lits}. On the contrary, images of multiple phases are required to detect and diagnose the liver lesions. Representatively, HCC warrants diagnostic imaging characteristics of arterial enhancement and portal or delayed washout as stated by major guidelines \cite{guideline}. Thus, the representational power of deep learning-based models \cite{christ:p,cascadedresnet,jointliver} is bounded by the data distribution itself. For example, specific variants of the lesion are difficult to see from the portal phase (Figure \ref{fig:figure1}). Therefore, a variety of hand-engineered data pre-processing techniques are required for deep learning with medical images.
\\\\
\indent Furthermore, from a clinical perspective, it is of practical value to detect lesion candidates by flagging them in real-time with a bounding box region of interest, which supports focused reading rather than pixel-wise segmentation, \cite{christ:p,cascadedresnet} which consumes a considerable amount of compute time. Considering the current drawbacks of the public datasets, we constructed a multi-phase detection CT dataset, which better reflects a real-world scenario of liver lesion diagnosis. While the segmentation dataset is more information-dense than the detection dataset, per-pixel labeling is less practical in terms of the scalability of the data, especially for medical images, which require skilled experts for clinically valid labelling. We show that the performance of our liver lesions detection model improves further when using multi-phase CT data.
\\\\
\indent We design an optimized version of the Single Shot MultiBox Detector (SSD) \cite{ssd}, a state-of-the-art deep learning-based object detection model. Our model incorporates grouped convolutions \cite{alexnet} for the multi-phase feature map. Our model successfully leverages richer information of the multi-phase CT data, while a naive application of the original model suffers from overfitting, which is where the model overly fits the training data and performs poorly on unobserved data.

\section{Multi-phase Data}
We constructed a 64 subject axial CT dataset, which contains four phases, for liver lesion detection. The dataset is approved by the international review board of Seoul National University Hospital. For image slices that contained lesions, we labeled such lesions in all phases with a rectangular bounding box. All the labels were determined by two expert radiologists. To enable the model to recognize information from the z-axis, we stacked three consecutive slices for each phase to create an input for the model. This resulted in a total of 619 data points, each of them having four phases aligned with the z-axis, and each of the phases having 3x512x512 image slices of the axial CT scan.
\\\\
\indent Since the volume of our dataset is much lower than the natural image datasets, the model unavoidably suffers more from overfitting, which is largely due to weakly-labeled ground truth bounding boxes. We labeled the lesions \textit{phase-wise}, rather than \textit{slice-wise}; for all slices that contain lesions in each phase, the coordinates of the bounding box are the same. While this method renders less burden on large-volume dataset construction, we get a skewed distribution of the ground truth, which hinders generalization of the trained model. To compensate for this limitation, we introduced a data augmentation for the ground truth, where we injected a uniform random noise to the bounding boxes to combat overfitting of the model while preserving the clinical validity of the labels. Formally, for each bounding box \(\textbf{y} = \{x_{min}, y_{min}, x_{max}, y_{max}\}\), we apply the following augmentation:
\begin{eqnarray} \label{eq:1}
\textbf{y}_{noise} = \textbf{y} \ \odot \ \textbf{z}, \ z_i \sim U(1-\alpha, 1+\alpha), 
\end{eqnarray}
where \(\odot\) is an element-wise multiplication, and \(\alpha > 0\) is set to a small value in order to preserve label information. We sample the noise on-the-fly while training the model.
\\\\
\indent We followed a contrast-enhancement pre-processing pipeline for the CT data in \cite{christ:p}. We excluded the pixels outside the Hounsfield Unit (HU) range [\(-\)100, 400] and normalized them to [0, 1] for the model to concentrate on the liver and exclude other organs. Since our dataset contains CT scans from several different vendors, we manually matched the HU bias of the vendors before pre-processing.

\section{Grouped Single Shot MultiBox Detector}
Here, we describe the SSD model and our modifications for the liver lesions detection task. In contrast to two-stage models \cite{fasterrcnn}, one-stage models \cite{yolo}, such as SSD, detect the object category and bounding box directly from the feature maps. One-stage models focus on the speed-accuracy trade-off \cite{huang2017speed}, where they aim to achieve a similar performance to two-stage models but with faster training and inference.
\begin{figure}[t]
	\centering
	\includegraphics[width=0.85\linewidth]{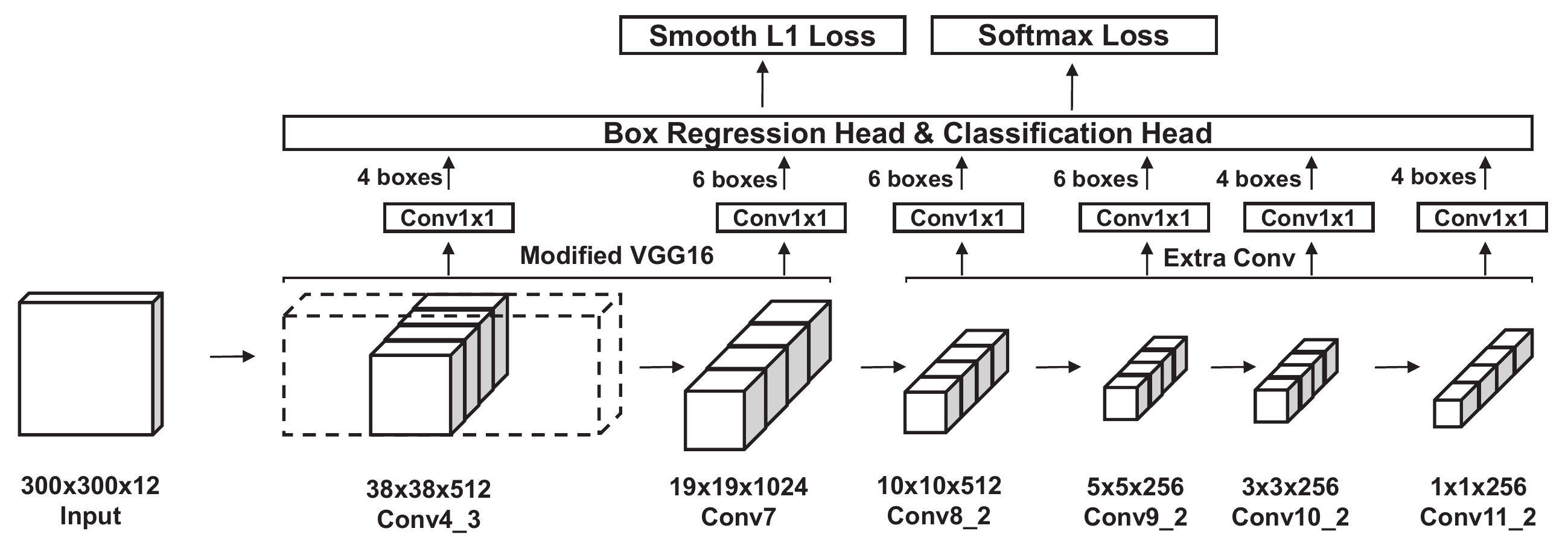}
    \caption{Schematic diagram of grouped Single Shot MultiBox Detector. Solid lines of the convolutional feature map at the bottom indicate grouped convolutions. Digits next to upper arrows from the feature maps indicate the number of default boxes for each grid of the feature map. Intermediate layers and batch normalization are omitted for visual clarity.}
    \label{fig:figure2}
\end{figure}
\\\\
\indent SSD is a one-stage model, which enables object detection at any scale by utilizing multi-scale convolutional feature maps (Figure \ref{fig:figure2}). SSD can use any arbitrary convolutional neural networks (CNNs) as base networks. The model attaches bounding box regression and object classification heads to several feature maps of the base networks. We use the modified VGG16 \cite{vgg} architecture as in the original model implementation to ensure a practical computational cost for training and inference. The loss term is a sum of the confidence loss from the classification head and the localization loss from the box regression head:
\begin{eqnarray}
	L(x, c, l, g) = \frac{1}{N}(L_{conf}(x, c) + L_{loc}(x, l, g)), 
\end{eqnarray}
where \(N\) is the number of matched (pre-defined) default boxes, \(x_{ij}^{p} = \{1, 0\}\) is an indicator for matching the \(i\)-th default box to the \(j\)-th ground truth box of category \(p\), \(L_{conf}\) is the softmax loss over class confidences \(c\) and \(L_{loc}\) is the smooth L1 loss between the predicted box \(l\) and the ground truth box \(g\).
\subsubsection{Grouped Convolutions}
Our custom liver lesions detection dataset consists of four phases, each of them having three continuous slices of image per data point, which corresponds to 12 ``channels'' for each input. We could apply the model naively by increasing the input channel of the first convolutional layer to 12. However, this renders the optimization of the model ill-posed, since the convolution filters need to learn a generalized feature representation from separate data distributions. This also runs the risk of exploiting a specific phase of the input, and not fully utilizing the rich information from the multi-phase input. Naive application of the model causes severe overfitting, which means the model fails to generalize to the unobserved validation dataset.
\\\\
\indent To this end, we designed the model to incorporate grouped convolutions. For each convolutional layer of the base networks, we applied convolution with separate filters for each phase by splitting the original filters, and concatenated the outputs to construct the feature map. Before sending the feature map to the heads, we applied additional 1x1 convolutions. This induces parts of the model to have separate roles, where the base networks learn to produce the best feature representation for each phase of the input, while the 1x1 convolutions act as a channel selector by fusing the grouped feature map \cite{nin,1x1conv} for robust detection.
\begin{figure}[t]
	\centering
    \includegraphics[width=\linewidth]{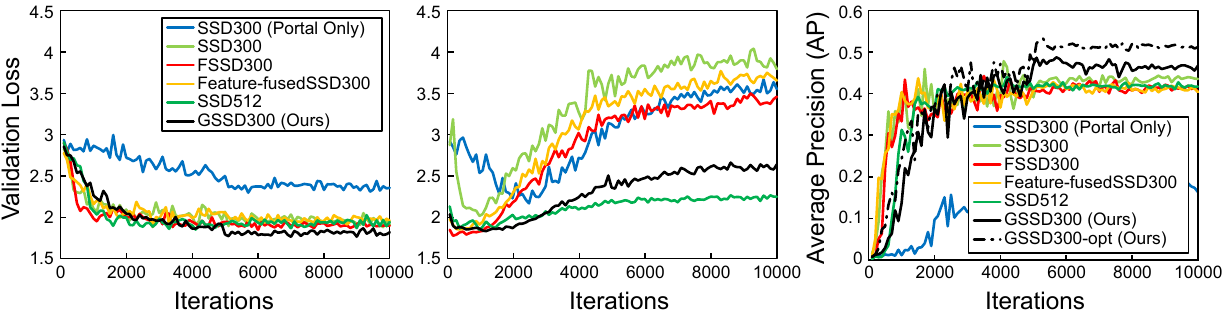}
    \caption{Performance comparison from the five-fold cross validation set. Left: Localization loss curves. Middle: Confidence loss curves. Right: Average precision scores. All models except GSSD300-opt used 1:1 OHNM (Table \ref{table:table1}).}
    \label{fig:figure3}
\end{figure}
\section{Experiments}
We trained the modified SSD models with our custom liver lesion detection dataset. For unbiased results, we employed five-fold cross validation. We applied all on-the-fly data augmentation techniques that were used in the original SSD implementation, but excluding hue and saturation randomization of the photometric distortion technique. We randomly cropped, mirrored, and scaled each input image (from 0.5 to 1.5). We trained the model over 10,000 iterations with a batch size of 16. We used a stochastic gradient descent optimizer with a learning rate of 0.0005, a momentum of 0.9, and a weight decay of 0.0005. We scheduled the learning rate adjustment with 1/10 scaling after 5,000 and 8,000 iterations for fine-tuning. We trained the models from scratch without pre-training, and initialized them using the Xavier method. We applied a batch normalization technique to the base networks for the grouped feature maps to have a normalized distribution of activations. We set the uniform random noise \(\alpha\) for the ground truth in Equation (\ref{eq:1}) to 0.01 for all experiments.
\begin{figure}[t]
	\centering
    \includegraphics[width=0.71\linewidth]{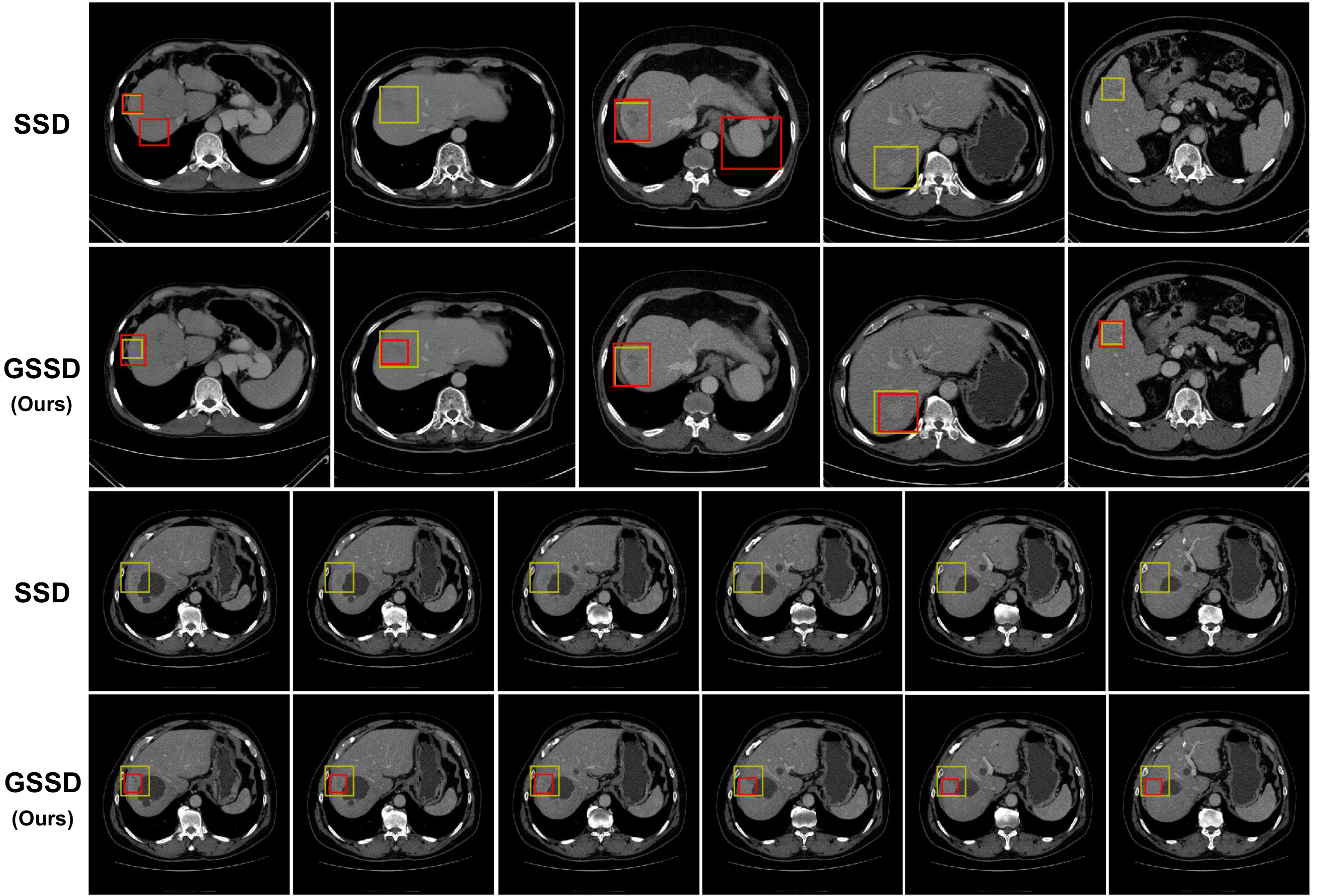}
    \caption{Qualitative results from the validation set with a confidence threshold of 0.3. Yellow: Ground truth box. Red: Model predictions. Portal images shown. Top: GSSD accurately detects lesions, whereas the original model contains false positives or fails to detect. Bottom: A case of continuous slices. GSSD successfully tracks the exact location of lesions even with the given weak ground truth, whereas SSD completely fails.}
    \label{fig:figure4}
\end{figure}
\\\\
\indent The performance definitively improved when using the multi-phase data. For comparison, the single-phase model received portal phase images copied four times as inputs. The model trained with only the portal phase data obviously underfitted (Figure \ref{fig:figure3}), since several variants of the ground truth lesions are barely visible from the portal CT images.
\\\\
\begin{table}[t]
\caption{Performance comparison of various configurations of SSD models. OHNM: The positive:negative ratio of Online Hard Negative Mining (OHNM) \cite{ssd}. 2xBase: Whether the model uses 2x feature maps in the base networks. \# 1x1 Conv: The number of layers for each feature map before sending to the heads. Best AP scores after 5,000 iterations reported.}
  \centering
  \scalebox{0.8}{
  \begin{tabular}{ l | c | c | c | c }
\toprule
  Model & {    OHNM    } & {    2xBase    } & {    \# 1x1 Conv    } & AP \\ 
  \midrule
  {SSD300 \cite{ssd} (Portal Only)    } & 1:1 & & & {    0.208    } \\
  \midrule
  SSD300 (in Figure \ref{fig:figure3}) & 1:1 & & & 0.444 \\
  SSD300 & 1:3 & & & \textbf{0.448} \\
  SSD300 & 1:1 & & 1 & 0.408 \\
  SSD512 (in Figure \ref{fig:figure3}) & 1:1 & & & 0.428 \\
  SSD512 & 1:3 & & & 0.433 \\
  FSSD300 \cite{fssd} & 1:1 & & & 0.432 \\
  Feature-fusedSSD300 \cite{featurefusedssd} & 1:1 & & & 0.437 \\
  \midrule
  GSSD300 & 1:1 & & & 0.445 \\
  GSSD300 & 1:1 & & 2 & 0.459 \\
  GSSD300 & 1:1 & \checkmark & 2 & 0.468 \\
  GSSD300 & 1:1 & \checkmark & 1 & \textbf{0.529} \\
  GSSD300 & 1:3 & \checkmark & 1 & 0.499 \\
  GSSD300 (in Figure \ref{fig:figure3}) & 1:1 & & 1 & 0.487 \\
  GSSD300-opt (in Figure \ref{fig:figure3}) & 1:3 & & 1 & \textbf{0.533} \\
 \bottomrule
  \end{tabular}
  }
  \label{table:table1}
\end{table}
\indent By significantly suppressing overfitting of the class confidence layers (Figure \ref{fig:figure3}), our grouped SSD (GSSD) outperformed the original model as well as recently proposed state-of-the-art variants (Table \ref{table:table1}) \cite{featurefusedssd,fssd}. Figure \ref{fig:figure4} demonstrates qualitative detection results. The best configuration achieved a 53.3\% average precision (AP) score (Table \ref{table:table1}). The model runs approximately 40 slices per second and can go through an entire volume of 100 slices in under three seconds on an NVIDIA Tesla P100 GPU. Note that the 1x1 convolutions play a key role as channel selectors. GSSD failed to perform well without the module. Stacking the 1x1 convolutions on top of the original model did not improve its performance, which proved that the combination of grouped convolutions and the channel selector module best harnesses the multi-phase data distribution.
\section{Discussion and Conclusions}
This study has shown that our optimized version of the SSD can successfully learn an unbiased feature representation from a weakly-labeled multi-phase CT dataset, which only requires phase-level ground truth bounding boxes. The system can detect liver lesions in a volumetric CT scan in near real-time, which provides practical merit for real-world clinical applications. The framework is also flexible, which gives it strong potential for pushing the accuracy of the model further by using more sophisticated CNNs as the base networks, such as ResNet \cite{resnet} and DenseNet \cite{densenet}.
\\\\
\indent We believe that the construction of large-scale detection datasets is a promising direction for fully leveraging the representational power of deep learning models from both machine learning and clinical perspectives. In future work, we plan to increase the size of the dataset to thousands of subjects, combined with a malignancy score label for the ground truth box for an end-to-end malignancy regression task.

\subsubsection{Acknowledgements.}
This work was supported by the National Research Foundation of Korea (NRF) grant funded by the Korea government (Ministry of Science and ICT) [2018R1A2B3001628], the Interdisciplinary Research Initiatives Program from College of Engineering and College of Medicine, Seoul National University (800-20170166), Samsung Research Funding Center of Samsung Electronics under Project Number SRFC-IT1601-05, and the Brain Korea 21 Plus Project in 2018.

%
%

%
%

\end{document}